# PROPAGATION OF BELIEF FUNCTIONS: A DISTRIBUTED APPROACH

by

Prakash P. Shenoy, Glenn Shafer and Khaled Mellouli

School of Business, University of Kansas, Lawrence, Kansas 66045-2003

## I. Abstract and Introduction

In this paper, we describe a scheme for propagating belief functions in certain kinds of trees using only local computations. This scheme generalizes the computational scheme proposed by Shafer and Logan[1] for diagnostic trees of the type studied by Gordon and Shortliffe[2,3] and the slightly more general scheme given by Shafer[4] for hierarchical evidence. It also generalizes the scheme proposed by Pearl[5] for Bayesian causal trees (see Shenoy and Shafer[6]).

Pearl's causal trees and Gordon and Shortliffe's diagnostic trees are both ways of breaking the evidence that bears on a large problem down into smaller items of evidence that bear on smaller parts of the problem so that these smaller problems can be dealt with one at a time. This localization of effort is often essential in order to make the process of probability judgment feasible, both for the person who is making probability judgments and for the machine that is combining them. The basic structure for our scheme is a type of tree that generalizes both Pearl's and Gordon and Shortliffe's trees. Trees of this general type permit localized computation in Pearl's sense. They are based on qualitative judgments of conditional independence.

We believe that the scheme we describe here will prove useful in expert systems. It is now clear that the successful propagation of probabilities or certainty factors in expert systems requires much more structure than can be provided in a pure production-system framework. Bayesian schemes, on the other hand, often make unrealistic demands for structure. The propagation of belief functions in trees and more general networks stands on a middle ground where some sensible and useful things can be done.

We would like to emphasize that the basic idea of local computation for propagating probabilities is due to Judea Pearl. It is a very innovative idea; we do not believe that it can be found in the Bayesian literature prior to Pearl's work. We see our contribution as extending the usefulness of Pearl's idea by generalizing it from Bayesian probabilities to belief functions.

In the next section, we give a brief introduction to belief functions. The notions of qualitative independence for partitions and a qualitative Markov tree are introduced in Section III. Finally, in Section IV, we describe a scheme for propagating belief functions in qualitative Markov trees.

## II. Belief Functions

Suppose $\Theta$ denotes a set of possible answers to some question, one and only one of which is correct. We call $\Theta$ a *frame of discernment*. A function Bel that assigns a degree of belief Bel(A) to every subset A of $\Theta$ is called a *belief function* if there is a random nonempty subset $S$ of $\Theta$ such that Bel(A) = Pr$[S \subseteq A]$ for all A.

Dempster's rule of combination is a rule for calculating a new belief function from two or more belief functions. Consider two random non-empty subsets $S_1$ and $S_2$. Suppose $S_1$ and $S_2$ are probabilistically independent, and suppose Pr$[S_1 \cap S_2 \neq \emptyset] > 0$. Let $S$ be a random non-empty subset that has the probability distribution of $S_1 \cap S_2$ conditional on $S_1 \cap S_2 \neq \emptyset$. If $Bel_1$ and $Bel_2$ are the belief functions corresponding to $S_1$ and $S_2$ then we denote the belief function corresponding to $S$ by $Bel_1 \oplus Bel_2$, and we call $Bel_1 \oplus Bel_2$ the *orthogonal sum* of $Bel_1$ and $Bel_2$. The rule for forming $Bel_1 \oplus Bel_2$ is called *Dempster's rule of combination*. Intuitively, $Bel_1 \oplus Bel_2$ represents the result of pooling the evidence represented by the separate belief functions whenever these items of evidence are independent.

A subset S of $\Theta$ is called a *focal element* of Bel if Pr$[S = S]$ is positive. In general, combination by Dempster's rule involves the intersection of focal elements. The focal elements for $Bel_1 \oplus ... \oplus Bel_n$ will consist of all non-empty intersections of the form $S_1 \cap ... \cap S_n$, where $S_i$ is a focal element of $Bel_i$. The computations involved in combining belief functions by Dempster's rule may become prohibitively complex when $\Theta$ is large since the number of subsets increases exponentially with the size of the frame. Hence it is important to exploit any special structure in the belief functions being combined that may help us reduce the computational burden.

One case where computational complexity of Dempster's rule can be reduced is the case where the belief functions being combined are "carried" by a partition $\wp$ of the frame $\Theta$. The complexity can be reduced in this case because $\wp$, which has fewer elements than $\Theta$, can in effect be used in the place of $\Theta$ when the computations are carried out.

250

Recall that a set $\wp$ of subsets of $\Theta$ is a *partition* of $\Theta$ if the sets in $\wp$ are all non-empty and disjoint, and their union is $\Theta$. Given a partition $\wp$ of $\Theta$, we denote by $\wp^*$ the set consisting of all unions of elements of $\wp$; $\wp^*$ is a field of subsets of $\Theta$ generated by $\wp$. We say that a belief function Bel over $\Theta$ is *carried* by $\wp$ if the random subset $S$ corresponding to Bel satisfies $\Pr[S \in \wp^*] = 1$. It is evident that a belief function Bel is carried by the partition $\wp$ generated by taking intersections of the belief function's focal elements. We can think of such a partition $\wp$ as a qualitative description of the belief function Bel and will refer to $\wp$ as the partition *associated* with Bel.

A partition $\wp$ of a frame $\Theta$ can itself be regarded as a frame. If Bel is a belief function on $\Theta$, then the *coarsening* of Bel to $\wp$ is the belief function $\text{Bel}_\wp$ on $\wp$ given by

$$\text{Bel}_\wp(\{P_1, ..., P_k\}) = \text{Bel}(P_1 \cup ... \cup P_k)$$ for every subset $\{P_1, ..., P_k\}$ of $\wp$. If Bel is a belief function on $\wp$, then the *vacuous extension* of Bel to $\Theta$ is the belief function $\text{Bel}^\Theta$ given by $\text{Bel}^\Theta(A) = \text{Bel}(\cup\{P \mid P \subseteq A, P \in \wp\})$. If a belief function is carried by $\wp$, then $\text{Bel}_\wp$ contains all the information about Bel. In fact, in this case, Bel can be recovered from $\text{Bel}_\wp$ by vacuous extension: $(\text{Bel}_\wp)^\Theta = \text{Bel}$. If $\wp_1$ and $\wp_2$ are two partitions, and Bel is a belief function on $\wp_1$, then the *projection* of Bel to $\wp_2$ is the result of vacuously extending Bel to $\Theta$ and then coarsening to $\wp_2$.

### III. Qualitative Markov Trees

The concept of conditional independence is familiar from probability theory, and it leads within probability theory to many other concepts, including Markov chains and Markov networks. In this section, we introduce a purely qualitative (non-probabilistic) concept of conditional independence and the corresponding concept of a qualitative Markov tree. Qualitative Markov trees are the setting for our computational scheme for propagating belief functions.

251

Let $\wp_1$ and $\wp_2$ be two distinct partitions. We say that $\wp_1$ is *coarser* than $\wp_2$ (or equivalently that $\wp_2$ is *finer* than $\wp_1$), written as $\wp_1 > \wp_2$, if for each $P_2 \in \wp_2$, there exists $P_1 \in \wp_1$ such that $P_1 \supseteq P_2$. We call $\wp_1$ a *coarsening* of $\wp_2$ and $\wp_2$ a *refinement* of $\wp_1$.

We write $\wp_1 \geq \wp_2$ to indicate that $\wp_1$ is coarser than or equal to $\wp_2$. The relation $\geq$ is a partial order and the set of all partitions is a lattice with respect to this partial order (Birkhoff[7]). The *coarsest common refinement* of $\wp_1, ..., \wp_n$ or the greatest lower bound of $\wp_1, ..., \wp_n$ with respect to $\geq$, denoted by $\wedge\{\wp_j \mid j = 1, ..., n\}$ or by $\wp_1 \wedge ... \wedge \wp_n$, is the partition $\{P_1 \cap ... \cap P_n \mid P_j \in \wp_j, \text{ for } j = 1, ..., n, \text{ and } P_1 \cap ... \cap P_n \neq \emptyset\}$.

We say that $\wp_1, ..., \wp_n$ are *qualitatively independent*, written as $[\wp_1, ..., \wp_n] \dashv$, if for any $P_j \in \wp_j$ for $j = 1, ..., n$, we have $P_1 \cap ... \cap P_n \neq \emptyset$. Furthermore, we say that $\wp_1, ..., \wp_n$ are *qualitatively conditionally independent* given $\wp$, written as $[\wp_1, ..., \wp_n] \dashv \wp$, if whenever we select $P \in \wp$, $P_i \in \wp_i$ for $i = 1, ..., n$ such that $P \cap P_i \neq \emptyset$ for $i = 1, ..., n$, then $P \cap P_1 \cap ... \cap P_n \neq \emptyset$. These definitions do not involve probabilities; just logical relations. But (stochastic) conditional independence for random variables does imply qualitative conditional independence for associated partitions (see Shafer, Shenoy and Mellouli[9] for details).

Qualitative conditional independence is important for belief functions because it is used in defining the circumstances under which we get the right answer when we implement Dempster's rule on a partition rather than on a finer frame (see Shafer[8], p. 177).

**Theorem 3.1** If $Bel_1$ and $Bel_2$ are carried by $\wp_1$ and $\wp_2$ respectively, and $[\wp_1, \wp_2] \dashv \wp$, then $(Bel_1 \oplus Bel_2)_\wp = (Bel_1)_\wp \oplus (Bel_2)_\wp$.

Another consequence of qualitative conditional independence is as follows (see Shafer,

252

Shenoy and Mellouli[9] for details and a proof of this result).

**Theorem 3.2** Suppose that $[\wp_1, \wp_2] \dashv \wp$. Let $Bel_2$ be carried by $\wp_2$. Then

$$(Bel_2)\wp_1 = ((Bel_2)\wp)\wp_1.$$

We now consider networks where the nodes represents partitions and the edges represent certain qualitative conditional independence restrictions on the partitions. An (undirected) *network* is a pair (J, E), where J, the *nodes* of the network, is a finite set, and E, the *edges* of the network, is a set of unordered pairs of distinct elements of J. We say that $i \in J$ and $j \in J$ are *adjacent* or *neighbors* if $(i, j) \in E$. A node is said to be a *leaf* node if it has exactly one neighbor. A network is a *tree* if it is connected and there are no cycles.

A *qualitative Markov network* for $\{\wp_j \mid j \in J\}$ is a network (J, E) such that given any three mutually disjoint subsets $J_1$, $J_2$, and $J_3$ of J, if $J_1$ and $J_2$ are *separated* by $J_3$ (in the sense that any path from a node in $J_1$ to a node in $J_2$ goes via some node in $J_3$), then

$$[\wedge\{\wp_j \mid j \in J_1\}, \wedge\{\wp_j \mid j \in J_2\}] \dashv \wedge\{\wp_j \mid j \in J_3\}.$$

If (J, E) is a qualitative markov network for $\{\wp_j \mid j \in J\}$ and the network (J, E) is a tree, then we say that (J, E) is a *qualitative Markov tree* for $\{\wp_j \mid j \in J\}$. A characterization of qualitative Markov trees is as follows (see Shafer, Shenoy and Mellouli[9] for a proof of this characterization).

**Theorem 3.3** Let $\{\wp_j \mid j \in J\}$ be a finite collection of partitions and let (J, E) be a tree. Given any node j in J, deletion of j from J and deletion of all edges incident to j from E results in a forest of k subtrees. Let the collection of nodes in the $i^{th}$ subtree be denoted by $\alpha_i(j)$. Then (J, E) is a qualitative Markov tree for $\{\wp_j \mid j \in J\}$ if and only if for every $j \in J$,

$$[\wedge\{\wp_i \mid i \in \alpha_1(j)\}, ..., \wedge\{\wp_i \mid i \in \alpha_k(j)\}] \dashv \wp_j.$$

253

## IV. Propagating Belief Functions in Qualitative Markov Trees

Suppose $T = (J, E)$ is a qualitative Markov tree for $\{\wp_i \mid i \in J\}$, and suppose that for every node i in J we have a belief function $Bel_i$ carried by $\wp_i$. We are interested in the orthogonal sum of all these belief functions, for which we use the symbol $Bel^T$:

$$Bel^T = \oplus\{Bel_i \mid i \in J\}.$$

We do not, however, need to know $Bel^T(A)$ for all subsets A of the frame $\Theta$. We need to know only $Bel^T(A)$ for certain A that are in the various fields $\wp_i^*$. This means that we will be satisfied if we can compute the coarsening $Bel^T \wp_i$ for every node i.

The coarsening $Bel^T \wp_i$ can in fact be computed efficiently by a simple recursive scheme that begins at the leaf nodes of T and moves towards node i while computing belief functions analogous to $Bel^T \wp_i$ for successively larger subtrees of T. This recursive scheme gains its efficiency from the fact that the computations it requires are local relative to the tree T. In place of a single global application of Dempster's rule, using $\Theta$ or $\wedge\{\wp_i \mid i \in J\}$ as our frame, we make many local applications of the rule, using the partitions $\wp_i$ as frames. Since the computational cost of the rule increases exponentially with the size of the frame, these numerous local applications can be inexpensive relative to a global application, provided the $\wp_i$ are all fairly small.

Given a subtree $U = \{J_U, E_U\}$ of T, let $Bel^U$ denote the orthogonal sum $\oplus\{Bel_i \mid i \in J_U\}$. Removal of node n (and all edges incident to node n) from T results in a set of subtrees, one for each neighbor k of n. Let $V_n$ denote the neighbors of n, and for each k in $V_n$, let $T_{k,n} = (J_{k,n}, E_{k,n})$ denote the subtree containing k (that results when n is removed from T). The basic relation that allows recursive computation of $Bel^T \wp_n$ is stated in Theorem 4.1 below.

**Theorem 4.1.** Let $T = \{J, E\}$ be a qualitative markov tree for $\{\wp_i \mid i \in J\}$ and let $Bel_i$ be carried by $\wp_i$ for each i in J. Then

$$Bel^T \wp_n = Bel_n \oplus (\oplus\{(Bel^{T_{k,n}}\wp_k)\wp_n \mid k \in V_n\}) \tag{4.1}$$

**Proof:** Since

$$Bel^T = Bel_n \oplus (\oplus\{Bel^{T_{k,n}} \mid k \in V_n\}),$$

$Bel_n$ is carried by $\wp_n$, $Bel^{T_{k,n}}$ is carried by $\wedge\{\wp_j \mid j \in J_{k,n}\}$, and

$$[\wedge\{\wp_j \mid j \in J_{k,n}\}]_{k \in V_n} \dashv \wp_n,$$

it follows from Theorem 3.1 that

$$Bel^T \wp_n = Bel_n \oplus (\oplus\{Bel^{T_{k,n}} \wp_n \mid k \in V_n\}).$$

Since

$$[\wp_n, \wedge\{\wp_j \mid j \in J_{k,n}\}] \dashv \wp_k$$

for every $k \in V_n$, it follows from Theorem 3.2 that

$$Bel^{T_{k,n}} \wp_n = (Bel^{T_{k,n}} \wp_k)\wp_n. \qquad \text{Q.E.D.}$$

The belief functions in the right hand side of (4.1), $Bel_n$ and $(Bel^{T_{k,n}}\wp_k)\wp_n$ for $k \in V_n$, are all carried by $\wp_n$, and hence their orthogonal sum can be computed using $\wp_n$ as a frame. The computation is recursive because $Bel^{T_{k,n}}\wp_k$ is the same type of object as $Bel^T \wp_n$, except that it is based on the smaller tree $T_{k,n}$. We need, of course, to get the recursion started; we need to be able to compute $Bel^U \wp_j$ when U is a tree containing only the node j, or perhaps j and some of its neighbors. But this is easy. If U consists of the single node j, then (4.1) tells us that

$$Bel^U \wp_j = Bel_j, \tag{4.2}$$

and if U consists of j and some of its neighbors, say in $V_j'$, then (4.1) tells us that

255

$$\mathrm{Bel}^U \wp_j = \mathrm{Bel}_j \oplus (\oplus\{(\mathrm{Bel}_k)\wp_j \mid k \in V_j'\}). \tag{4.3}$$

We can take either (4.2) or (4.3) as the starting point of the recursion.

In order to see more clearly how to direct the recursion, let us introduce some further notation. Given two neighboring nodes i and j in the tree T, set

$$\mathrm{Bel}_{j \to i} = (\mathrm{Bel}^{T_{j,i}} \wp_j) \wp_i$$

where, as noted before, $T_{j,i}$ denotes the subtree containing j that results when i is removed from T. With this notation, (4.1) can be written as

$$\mathrm{Bel}^T \wp_n = \mathrm{Bel}_n \oplus (\oplus\{\mathrm{Bel}_{k \to n} \mid k \in V_n\}). \tag{4.4}$$

Moreover, Theorem 4.1 applied to $T_{j,i}$ tells us that

$$\mathrm{Bel}_{j \to i} = \left(\mathrm{Bel}_j \oplus (\oplus\{\mathrm{Bel}_{k \to j} \mid k \in (V_j - \{i\})\})\right) \wp_i \tag{4.5}$$

for any neighboring nodes i and j. If j is a leaf node and i is its only neighbor, then the set $V_j - \{i\}$ is empty, and then (4.5) says simply that $\mathrm{Bel}_{j \to i} = (\mathrm{Bel}_j) \wp_i$.

Formulae (4.4) and (4.5) suggest a very simple way to program our recursive computations of $\mathrm{Bel}^T \wp_i$ in a forward chaining production system. We begin with a working memory that contains $\mathrm{Bel}_i$ for each node i in J, and we use just two rules:

**Rule 1:**

*If* $j \in J$, $i \in V_j$, $\mathrm{Bel}_{k \to j}$ is present in working memory for every k in $V_j - \{i\}$, and $\mathrm{Bel}_j$ is present in working memory,

*then* use (4.5) to compute $\mathrm{Bel}_{j \to i}$ and place it in working memory.

**Rule 2:**

*If* $i \in J$, $\mathrm{Bel}_{k \to i}$ is present in working memory for every k in $V_i$, and $\mathrm{Bel}_i$ is present in working memory,

*then* use (4.4) to compute $\mathrm{Bel}^T \wp_i$, and then print it.

256

Notice that Rule 1 will fire initially only for leaf nodes, since initially no $Bel_{k \to j}$ are in working memory. Rule 1 will eventually fire in both directions for every edge (i, j) producing both $Bel_{j \to i}$ and $Bel_{i \to j}$. We assume that repetitions of these firings are prevented by a refractory principle that prevents a rule from firing again for the same instantiation of the antecedent. Rule 2 will eventually fire for every i. Thus the total number of firings is equal to $2(|J| - 1) + |J| = 3|J| - 2$.

The potential efficiency of this computational scheme is enhanced by the fact that many of the applications of Dempster's rule on different $\wp_i$ can be carried out in parallel. We can make this potential parallelism graphic by imagining that a separate processor is assigned to each $\wp_i$. The processor assigned to $\wp_i$ computes $Bel^T \wp_i$ and $Bel_{i \to k}$ using (4.4) and (4.5) respectively. This means that it combines belief functions using $\wp_i$ as a frame. It also projects belief functions from $\wp_i$ to $\wp_k$, where k is a neighbor of i.

Since the processor assigned to $\wp_i$ communicates directly with the processor devoted to $\wp_k$ only when k is a neighbor of i, the Markov tree itself can be thought of as a picture of the architecture of the parallel machine; the nodes are processors and the links are communication lines. In this parallel machine, the "working memory" of the production system implementation is replaced by local memory registers at the links. We may assume that every link, there are two sets of memory registers -- one for communication in each direction. Thus at the link between i and k, say, there will be one set of registers where i writes $Bel_{i \to k}$ for k to read, and another where k writes $Bel_{k \to i}$ for i to read. Each processor i also has an input register, where $Bel_i$ is written from outside the machine, and an output register, where it writes $Bel^T \wp_i$. Figure 4.1 shows a typical processor, with three neighbors.

We may assume that the processor at i begins work on the computations it is authorized to perform as soon as it receives the necessary inputs. In other words, it computes $Bel_{i \to j}$ as soon

257

as it receives $Bel_i$ and $Bel_{k \to i}$ for all $k \in V_i - \{j\}$, and it computes $Bel^T \wp_i$ as soon as it receives $Bel_i$ and $Bel_{k \to i}$ for all $k \in V_i$. If we further assume that the processor does not repeat computations for the same inputs, and if we input all the $Bel_i$ before turning the processors on, then our parallel machine will operate in fundamentally the same way as the production system we described above.

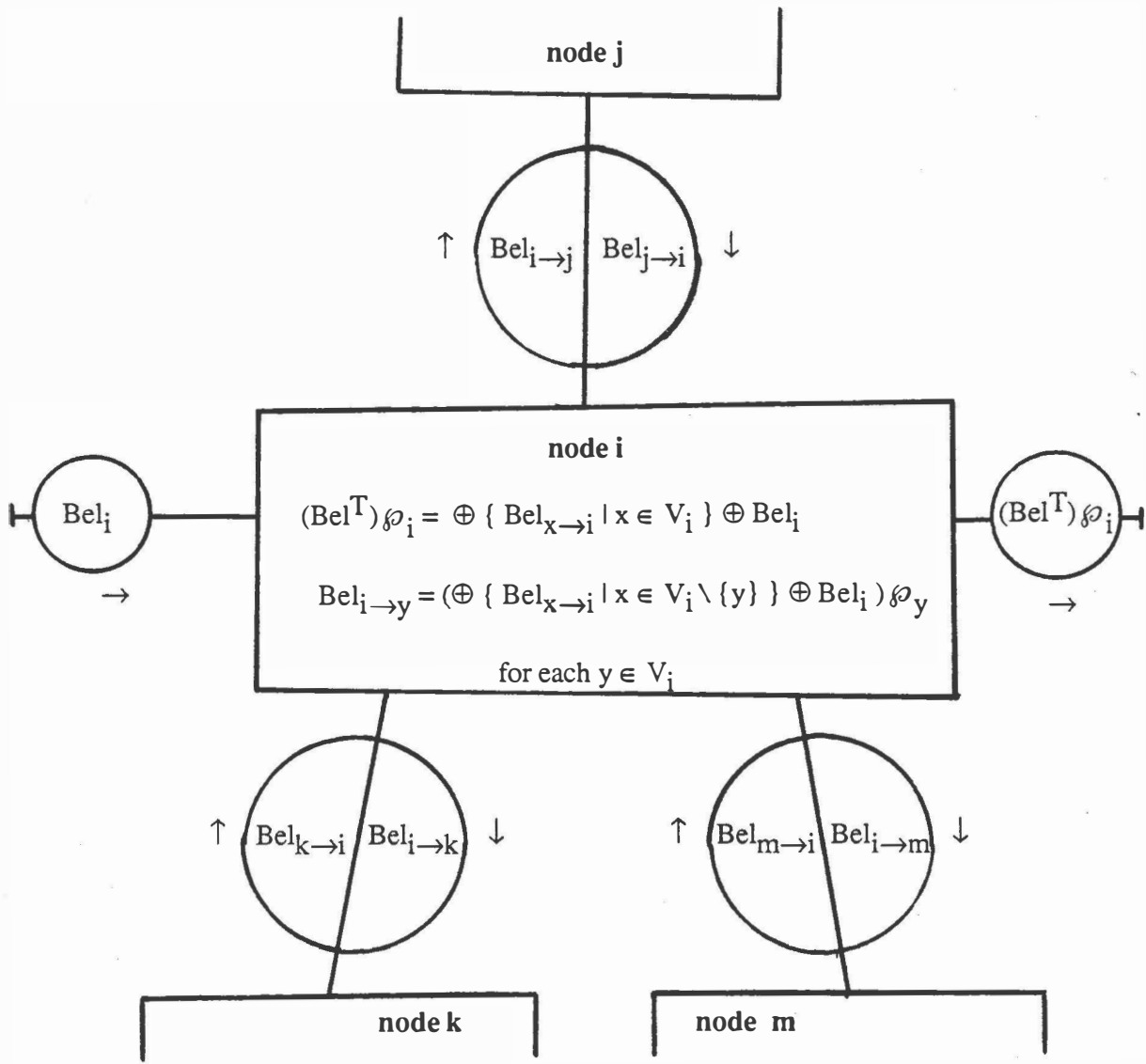

**Figure 4.1** A typical node processor (with three neighbors).

258

The parallel machine could also be operated in a more dynamic way. Instead of entering all $Bel_i$ before starting the computations, the $Bel_i$ could be entered at any time. Initially, to get the computations started, we let all the belief functions $Bel_i$ to be vacuous. Then as we accumulate independent pieces of evidence, we enter these (non-vacuous) belief functions reprsenting the evidence at the appropriate nodes in the tree. Note that if we have two or more independent pieces of evidence that is represented by belief functions carried by the same node $\wp_i$, and these belief functions are all entered at $\wp_i$, then the processor at $\wp_i$ combines all the belief functions input to it to form a belief function $Bel_i$. Also note that the refractory principle does not prevent the two rules from firing again for the same edges and nodes if the instantiation is different (as a result of entering a new belief function at a node).

## V. Conclusion

The scheme described above is not an algorithm. It does not specify how the coarsenings from one partition to its neighbors are to be carried out, since the most efficient way to do this will depend on the particular nature of the relations between these partitions. The details of the implementation of Dempster's rule at the level of each partition may also depend on the nature of the belief functions being combined there. The general scheme is useful, however, because of its conceptual clarity and its unifying role. In particular, it unifies two computational schemes that had previously seemed rather disparate: Pearl's scheme for propagation of probabilities in Bayesian causal trees (Pearl[5]) and Shafer and Logan's scheme for combining belief functions in diagnostic trees (Shafer and Logan[1]). Both these schemes are special cases of the general scheme for propagation in qualitative Markov trees, and they derive most of their computational power from this fact though they also exploit special features of the problem they solve. Pearl's scheme derives some computational power from the simplicity of Bayesian probability measures relative to general belief functions, and Shafer and Logan's scheme derives some computational power from Barnett's technique (Barnett[10]) which it is able to exploit because the belief functions being combined are "simple support functions" (i.e., have atmost two focal elements one of which is the frame $\Theta$). A comparison of Pearl's and Shafer and Logan's schemes with the general scheme presented here is sketched in Shenoy and Shafer[6].

## VI. Acknowledgements

Research for this paper has been partially supported by NSF grant IST-8405210 and grant P. O. S-25780 from Harvard University.